\documentclass[letterpaper]{article}
\usepackage{aaai20}
\usepackage{times}
\usepackage{helvet}
\usepackage{courier}
\usepackage[hyphens]{url} 
\usepackage{graphicx}

\usepackage{enumitem}
\usepackage{multirow}

\usepackage{graphicx}
\title{Unsupervised Key-phrase Extraction and Clustering for Classification Scheme in Scientific Publications}

\author{        
        Xiajing Li,\textsuperscript{\rm 1, \rm 2}
        Marios Daoutis,\textsuperscript{\rm 1}\\ 
    \textsuperscript{\rm 1}Ericsson AB, \\
    \textsuperscript{\rm 2}Department of Linguistic and Philology, Uppsala University\\
    xiajing.li@ericsson.com, marios.daoutis@ericsson.com
}

\begin{document}
\maketitle

\begin{abstract}

A Systematic Review of a research domain provides a way to understand and structure the state-of-art of a particular research area. Extensive reading and intensive filtering of large volumes of publications are required during that process, while almost exclusively performed by human experts. Automating sub-tasks from the well defined Systematic Mapping (SM) and Systematic Review (SR) methodologies is not well explored in the literature, despite recent advances in natural language processing techniques. Typical challenges evolve around the inherent gaps in the semantic understanding of text and the lack of domain knowledge necessary to fill-in that gap. In this paper, we investigate possible ways of automating common sub-tasks of the SM/SR process, i.e., extracting keywords and key-phrases from scientific documents using unsupervised methods, which are then used as a basis to construct the so-called classification scheme using semantic clustering techniques. Specifically, we explore the effect of ensemble scores in key-phrase extraction, semantic network-based word embeddings as well as how clustering can be used to group related key-phrases. We conducted an evaluation on a dataset from publications on the domain of ``\textit{Explainable AI}'' which we constructed from standard, publicly available digital libraries and sets of indexing terms (keywords). Results show that ensemble ranking score does improve the key-phrase extraction performance. Semantic network-based word embeddings (ConceptNet) has similar performance as contextualized word embeddings, while the former is more efficient than the latter. Finally, semantic term clustering can group similar terms, which can be suitable for classification schemes.
\end{abstract}

\section{Introduction}

Systematic Mapping (SM) and Systematic Review (SR) studies are standard methods for capturing the state-of-art of a particular research field in a structured and organised way, while at the same time provide significant insights and knowledge around that research area \cite{mapping}. Traditionally, these methods are performed manually by human experts and researchers. With a growing number of publications in recent years as well as the literature expansion in novel areas, the systematic mapping procedure of such volumes of scientific documents becomes quite challenging and time-consuming \cite{BarriersSLR}.

In the classical systematic mapping procedure, \textit{keyword extraction} \& \textit{classification scheme} are two essential steps that help in classifying papers in different perspectives while producing a group of categories from, typically, manual keywording and grouping of the descriptive terms. First, terms extracted by intensively reading papers should be common in regard each source document as well as the research domain. Existing keywords and key-phrase extraction systems are usually independent, concerning downstream tasks and types of documents. For document types, such as web pages and social media documents, short and concise keywords are required, while multi-word expressions (key-phrases) are more common in scientific publications.

In this work we explore methods that can leverage the identified and automatically extracted keywords for producing a classification scheme for the research domain of interest. Furthermore, we evaluate methods suitable for extracting representative (as an attribute of each document) and highly relevant (to a target research domain) keywords drawn from the summary (abstract) of scientific publications. We are interested in getting keywords and key-phrases that are precise yet informative as domain concepts or terminologies.

Hence, we attempt to address whether automated key-phrase extraction methods and term clustering techniques can adequately extract and identify useful information, comparable to how they are performed in the context of SM \& SR. More specifically, we explore the effect of ensemble score measures in key-phrase extraction (\textbf{Q1}), the effect of semantic network-based word embedding techniques in embedding representation of phrase semantics (\textbf{Q2}), as well as the effect of clustering for grouping semantically related key-phrases (\textbf{Q3}). Our code and data will be publicly available at: https://github.com/xiajing10/akec.

\section{Related Work}

With an increasing number of research publications, especially in artificial intelligence, current systematic mapping underlying procedures are time-consuming. The survey from \citeauthor{BarriersSLR} discusses the barriers of manual work in the systematic literature review process, especially in the context of paper selection and data extraction \cite{BarriersSLR}. Recent text-mining algorithms and NLP techniques can become particularly useful for automating (parts of) this manual work within the systematic mapping studies procedure. Several studies have investigated various techniques to automate one or more sub-steps, such as paper selection \cite{towards-slr-ml}. However, we find that very few  related works focus on automating the process of \textit{keywording} and \textit{categorization} steps, which presume background knowledge from domain experts. Extracted keywords have to encode salient (essential and relevant) text features and the aspect of human readability (as concepts). Then, when grouping sets of keywords into different categories, human experts have an inherent ability to understand the definition, background knowledge, and semantic relatedness of keywords. 

Keyword extraction generates highly representative and relevant information from unstructured text, used as features in many downstream tasks, such as summarization, clustering, knowledge graph generation, and taxonomies. Unsupervised systems typically apply scoring and ranking methods on candidate words. TF-IDF is a simple but effective scoring mechanism. Graph-based methods (e.g., TextRank \cite{mihalcea-tarau-2004-textrank}) rank the importance of words based on word co-occurrence graph, which has shown its effectiveness independently of domain and language. Semantic information of words is rarely used in early methods, as it is usually difficult to measure. Word embedding techniques provide a means to measure such semantic similarity. Semantic similarity between each candidate and its source document can be calculated by \textit{cosine similarity} of their embedding representation. \citeauthor{papagiannopoulou2020review} utilize averaging GloVe word embedding as phrase vector and ``theme vector" \cite{papagiannopoulou2020review}. \citeauthor{BennaniSmires2018EmbedRankUK} applies Doc2Vec and Sent2Vec for document representation and phrase representation \cite{BennaniSmires2018EmbedRankUK}. \citeauthor{SIFrank} combined various contextualized word embedding methods with SIF weighted sentence embedding model \cite{SIFrank}. In this paper, we further explore the performance of semantic network based word embeddings building on the work of SIFRank.

A pre-existing classification scheme, typically, does not always fit more than one particular research domain. Updating or generating a new classification scheme from selected papers is widely applied in most cases, with help from text-mining techniques. \citeauthor{terko2019neurips} conducted conference paper classification using traditional machine learning methods, with labels generated from topic modeling \cite{terko2019neurips}. \citeauthor{kim2019research} applied $k$-means as an unsupervised clustering method for creating the classification scheme at a document-level, during which they extracted features from topic models, abstracts and author-given keywords, followed by TF-IDF vectorization and document clustering \cite{kim2019research}. Different from categories in systematic mapping studies, document clustering is single-faceted, where each article is assigned to only one category. \citeauthor{Osborne2019ReducingTE} proposed their semi-supervised system for mapping studies, which starts with ontology learning over large scholarly datasets, then refines the ontology with the help of domain experts, and finally use knowledge bases to select and classify the primary studies automatically \cite{Osborne2019ReducingTE}. Their classification scheme is generated by selecting several ontologies from author-given keywords as categories and identified equivalent ontologies (based on relations learned in ontology learning) as they appeared in abstracts, keywords, and titles. Their approach shows higher precision compared to TF-IDF. However, it relies on an extensive, extracted database of ontologies of author-given keywords, which are sometimes missing in attributes. Unlike the methods discussed above, our method is inspired by taxonomy generation by term clustering, which focuses on grouping words/terms similarity based on their representation. Using taxonomy as a classification scheme would be more suitable in immature or evolving domains than classification with fixed classes \cite{taxo-review-usman}. \citeauthor{liu2012automatic} construct taxonomy from keywords using hierarchical clustering \cite{liu2012automatic}. \citeauthor{zhang2018taxogen} generates taxonomy using spherical $k$-means to cluster terms extracted from a large-scale set of publications from the domain of computer science, with word embeddings learned from the text. Considering that a large corpus is not always obtainable, we first apply keyword extraction to extract terms \cite{zhang2018taxogen}.


\section{Methodology}

\begin{figure}[htb]
    \centering
    \includegraphics[width = 0.45\textwidth]{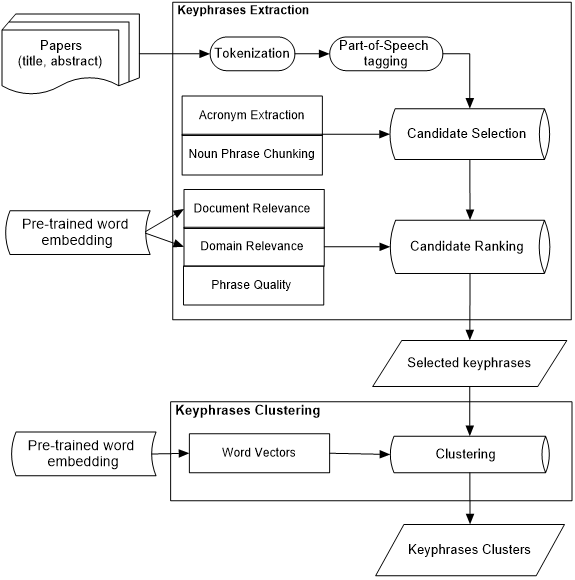}
    \caption{Framework of proposed automation method.}
    \label{fig:framwork}
\end{figure}

Our automation method follows the pipeline of classification scheme generation \cite{SMS-keywording}. It is composed of two modules: (1) key-phrase extraction from titles and abstracts; (2) term clustering to identify key-phrases categories. Our system's overall framework is shown in Fig. \ref{fig:framwork}, leveraging a semantic similarity measure and external knowledge from pre-trained word embedding. 

\subsection{Key-phrase Extraction}

The key-phrase extraction module is built based on SIFRank \cite{SIFrank}, a state-of-art embedding-based method, whose pipeline consists of (1) candidates selection by noun phrase chunking and (2) candidates ranking by candidate-document cosine similarity. We use the SIFRank score to measure document relevance, together with two other scoring functions for measuring domain relevance and phrase quality. The three scores are combined for candidate key-phrases ranking.

\paragraph{Document relevance score} One keyword of a single document should have a strong connection with this document. Semantic distance with word embedding is based on the principle that as closer a candidate vector is to the document vector, the closest the distance is in regard to their meanings. The effectiveness of the semantic distance measure has been previously evaluated in benchmark datasets \cite{BennaniSmires2018EmbedRankUK}. SIFRank\cite{SIFrank} reaches state-of-art performance in key-phrase extraction for short documents, while utilizing auto-regressive pre-trained Language Model ELMo to produce word embedding and SIF (Smooth Inverse Frequency) \cite{arora2017sif} to generate unsupervised sentence embedding. In scientific publications, representative key-phrases frequently appear in titles. Each candidate's final document relevance score is the original score weighted according to the candidates that appear in the title. The weight is defined by the length of the tokens of candidate phrases.

\paragraph{Domain Relevance Score} Finding domain-specific terms has been a challenge for novel domains with fewer related resources (publications). Terms with high frequency in domain-specific corpus and low frequency in other domains can be considered domain-specific terms. In contrast, without a domain-specific corpus, dictionary-based validation can help to improve finding representative terms. Structured semantic resources (e.g., WordNet) can help in utilizing semantic relations, such as groups of synonyms or topic-based clusters, assuming that related terms are more likely to be critical than isolated ones \cite{firoozeh2020keywordreview}. In the general systematic mapping studies process, glossary dictionary and domain seed key-phrases are provided with the help of human experts. Here we collect our domain glossary terms from open knowledge graph databases: (1) artificial intelligence knowledge graph \cite{aikg} using terms with direct link connections to ``artificial intelligence"; (2) machine learning taxonomy from Aminer \cite{tang2016aminer}. Semantic similarity between candidates and glossary terms are calculated for relevance scoring.  Detailed steps are described below:

\begin{quote}
\begin{enumerate}[label=\textbf{{Step \arabic{enumi}}.},leftmargin=*]
    \item Candidate key-phrases and domain glossaries are transformed by pre-trained word embedding.
    \item For each candidate phrase, scosine similarity is calculated between itself and each domain glossary.
    \item Domain relevance score of one candidate phrase is the average of top N ($50$\% in our experiment) highest similarity scores.
\end{enumerate}
\end{quote}

\paragraph{Phrase Quality Score} In scientific documents, high-quality phrases are usually multi-word expressions or uni-grams as an acronym, representing common or newly defined scientific concepts. Therefore, our method considers this fact and defines the quality score of a term according to length penalty, point-wise mutual information (PMI), left-right information entropy strategy, and acronym information. The length penalty aims to reduce the score of uni-grams and long phrases. Based on the analysis of the scientific documents dataset, the majority of gold key-phrases are bi-grams and tri-grams. Hence, we added length penalty to multi-word expression $t$ that contains more than three words as $length\_score (t) = - 0.5 * \left \| length(t) - 3 \right \|$. However, acronyms are extensively used as a shorter format (mostly uni-grams) of long scientific terms. Since acronym usually refers to a specific terminology or scientific concept in the document, it is a good indicator of whether the term is important or not. Therefore, the length penalty does not apply to uni-grams that are identified as acronyms. The well-known PMI and entropy strategies are used to extract multi-word expressions that co-occur frequently and contain a collective meaning. Generally, a high PMI score indicates a high probability of co-occurrence. We calculated the minimum PMI score among all two segments of the expression for expressions that contain more than two words. For example, the score of \textit{``explainable artificial intelligence''} is equal to the minimum score of PMI(x=\textit{explainable machine}, y=learning) and PMI(x=explainable, y=\textit{machine learning}). Left-right information entropy (Eq. \ref{eq:entropy}) shows the variety of word context of a candidate phrase and adjacent words will be widely distributed if the string (candidate phrase) is meaningful, and they will be localized if the string is a sub-string of a meaningful string \cite{shimohata1997retrieving}.

\begin{equation}
\label{eq:entropy}
  H(t)=-\sum_{w_{i} \in wl} p{(w_{i}|t)}\,log_2\,p{(w_{i}|t)}\
\end{equation}
where \textit{wl} represent the list of adjacent words of candidate phrase \textit{t}. Both left and right sides of phrase \textit{t} is calculated and the lower one is selected as the final information entropy score. 

In detail, the quality score of a candidate term $t$ is the sum of \emph{PMI-entropy score} and \emph{length penalty} . To weaken PMI's bias towards low frequency words, we filter out candidate terms with low PMI score (threshold at PMI = 2 in our experiments) and use the normalized entropy score of the rest candidate terms as \emph{PMI-entropy score}. 


\subsection{Key-phrase Clustering} Clustering aims to identify distinct groups in a dataset and assign a group label to each data point. This module focuses on clustering key-phrase based on their semantic similarity (cosine similarity of their embedding representation). For this module we tested two clustering algorithms: spherical $k$-means and hierarchical agglomerative clustering. As bottom-up clustering, agglomerative clustering starts with each data point as an individual cluster and then merges sub-clusters into one super-cluster based on a certain distance threshold. Spherical $k$-means is $k$-means on a unit hyper-sphere, where (1) all vectors are normalized to unit-length and (2) objective function is to minimize cosine distance between vectors. Studies have found the effectiveness of cosine similarity in quantifying the semantic similarities between high dimensional data such as word embedding, as the direction of a vector is more important than the magnitude \cite{Strehl00impactof}. Comparing to standard $k$-means, spherical of $k$-means matches the distinct nature of cosine similarity measure in words embedding high dimensional space. \citeauthor{zhang2018taxogen} illustrates that when using spherical $k$-means for topic detection, the center direction acts as a semantic focus on the unit sphere, and the member terms of that topic fall around the center direction to represent a coherent semantic meaning \cite{zhang2018taxogen}. 

\section{Experimental Evaluation}

This section presents our experimental evaluation setup for our proposed automation approach. We aim at answering the following questions:
\begin{itemize}
    \item \textbf{Q1:} Can our ensemble scoring measure improve performance in domain-specific key-phrase extraction?
    \item \textbf{Q2:} How does semantic network based word embedding techniques (ConceptNet) perform in embedding representation of phrase semantics?
    \item \textbf{Q3:} Does the clustering method group semantically related key-phrases for identifying categories?
\end{itemize}

\subsection{Data}

Data collection determines the quality and relevance of the further steps of systematic mapping studies. As keywording follows after the step of paper selection, we assume that the selected input articles under consideration for our framework are considered to be already \textit{in-domain}. However, common benchmark datasets for key-phrase extraction from scientific articles do not focus on a specific research domain. We collected a set of scientific articles from \textbf{IEEE Xplore} under the domain of ``\textit{Explainable Artificial Intelligence}''. In total, 286 scientific publications were extracted together with their meta-data attributes, which we name \textit{XAI dataset}. ``Title'' and ``abstract'' of each article were combined as input text. Also, IEEE Xplore provides INSPEC indexing terms assigned by human experts to represent a publication's content. For the evaluation of the key-phrase extraction, we use the ``INSPEC Non-Controlled Indexing terms'' attribute as a gold standard, as its terms are primarily emerge from text.

\begin{table}[htb!]
\centering
\resizebox{0.45\textwidth}{!}{
\begin{tabular}{|c|c|c|c|c|}
\hline
                   & total & in text & \begin{tabular}[c]{@{}c@{}}average \#nums \\ of tokens\end{tabular} & \begin{tabular}[c]{@{}c@{}}average \#count \\ per paper\end{tabular} \\ \hline
Non-Controlled terms & 3200  & 88.84\%& 2.6181                                                             & 11.1888                                                              \\ \hline
Controlled terms & 1536  & 20.73\%        & 2.1978                                                             & 4.7727                                                               \\ \hline
\end{tabular}}
\caption{Comparative analysis of \textit{Non-Controlled indexing terms} and \textit{Controlled inxdexing terms}.}
\label{table:con-uncon}
\end{table}

\subsection{Implementation and Tools}

\paragraph{Pre-processing} The title and abstract of each document are concatenated as input text. Initial experiments on candidates selection recall found that lowercase and punctuation removal would affect acronym extraction, tokenization, and noun phrase chunking. Also, noun phrases with dash tag will lead to a low recall of correct candidates. Thus, we remove applied dash tags and use an extended set of common stopwords\footnote{Stopwords list from https://www.ranks.nl/stopwords}.

\paragraph{Candidate Selection}

Candidate selection is built under the framework of  SIFRank\footnote{https://github.com/sunyilgdx/SIFRank} model, where tokenizer and POS tagger have been changed to SpaCy. Noun phrase pattern (defined as in Eq. \ref{eq:np}) is captured by regular expressions and parsed into constituency tree for pattern matching. 

\begin{equation}\label{eq:np}
    <NN.*|JJ>*<NN.*>
\end{equation}

Acronym Extraction is implemented directly using build-in function in ScispaCy. Considering that acronym are case-sensitive, we implemented acronym extraction before pre-processing.

\paragraph{Candidate Ranking}

Details of the candidate scoring process are illustrated above. The latest version of pre-trained ConceptNet numberbatch (ConceptNet Numberbatch 19.08, English version) is used as pre-trained word embedding for embedding representation. Our domain glossary terms are selected from the open resources knowledge graph database: (1) artificial intelligence knowledge graph\footnote{\url{http://scholkg.kmi.open.ac.uk/}} \cite{aikg}: terms with direct link connection with the term ``artificial intelligence" are extracted; (2) machine learning taxonomy from Aminer\footnote{\url{https://www.aminer.cn/data}} \cite{tang2016aminer}.

\paragraph{Selection of Key-phrases}

Before moving forward to the clustering module, post-processing controls the quality of the extracted key-phrases to match the use case. We defined a few rule-based steps for post-processing:

\begin{enumerate}
    \item Lemmatize key-phrases to remove redundant key-phrase due to language inflection. The higher score between the two will be assigned.
    \item Average rank of key-phrases among documents. Key-phrases ranked above 15 are selected.
    \item Replace key-phrase identified as an acronym by its original definition in text. 
    \item Remove last 20\% key-phrases based on TF-IDF scores.
\end{enumerate}

\paragraph{Clustering Algorithms} Clustering module is built on scikit-learn \cite{scikit-learn} and spherecluster\footnote{https://pypi.org/project/spherecluster/}. Before clustering, each term will be transformed to embedding representation from ConceptNet Numberbatch. We first explore the optimal $k$ in range from 5 to 100 clusters. 

\begin{table*}[htb]
\resizebox{\textwidth}{!}{
\begin{tabular}{|c|c|c|c|c|c|c|c|c|c|c|}
\hline
                                    &                  & \multicolumn{3}{c|}{Top5} & \multicolumn{3}{c|}{Top10} & \multicolumn{3}{c|}{Top15} \\ \hline
                                    &                  & P       & R      & F1     & P       & R       & F1     & P       & R       & F1     \\ \hline
\multirow{2}{*}{TextRank}           & Baseline         & 0.4986  & 0.2228 & 0.3080 & 0.4411  & 0.3941  & 0.4162 & 0.3791  & 0.5066  & 0.4337 \\  
                                    & Combined Scoring & 0.5203  & 0.2325 & 0.3214 & 0.4627  & 0.4134  & 0.4367 & 0.3793  & 0.5069  & 0.4339 \\ \hline
\multirow{2}{*}{SIFRank-ELMo}       & Baseline         & 0.5105  & 0.2281 & 0.3153 & 0.4327  & 0.3866  & 0.4083 & 0.3803  & 0.5072  & 0.4347 \\ 
                                    & Combined Scoring & 0.5469  & 0.2444 & 0.3378 & 0.4834  & 0.4319  & 0.4562 & 0.4152  & 0.5538  & 0.4746 \\ \hline
\multirow{2}{*}{SIFRank-Bert}       & Baseline         & 0.5266  & 0.2353 & 0.3253 & 0.4418  & 0.3947  & 0.4169 & 0.3796  & 0.5063  & 0.4339 \\  
                                    & Combined Scoring & 0.5147  & 0.2300 & 0.3179 & 0.4530  & 0.4047  & 0.4275 & 0.3993  & 0.5325  & 0.4563 \\ \hline
\multirow{2}{*}{SIFRank-ConceptNet} & Baseline         & 0.5049  & 0.2256 & 0.3119 & 0.4257  & 0.3803  & 0.4017 & 0.3679  & 0.4906  & 0.4205 \\  
                                    & Combined Scoring & 0.5510  & 0.2463 & 0.3404 & 0.4774  & 0.4266  & 0.4506 & 0.4103  & 0.5471  & 0.4689 \\ \hline
\end{tabular}}
\caption{Comparison of key-phrase extraction results from ensemble methods with three base models.}\label{table:ensemble}
\end{table*} 

\subsection{Evaluation Metrics}

We evaluated our automation method using two criteria: reliability of extracted key-phrases and the quality of generated categories based on key-phrases. Evaluation is conducted separately on two modules. Evaluation of ranked key-phrase list used traditional statistical measures of Precision, Recall, and F1-score with the labeled gold standard. Morphological variants of phrases have been removed before evaluation. Evaluation of semantic term clustering lacked a ground truth classification scheme. We utilized an internal evaluation metric of the silhouette coefficient score to measure how well the cluster is separated.

\section{Results}

To investigate the feasibility of our proposed automation method, we conducted experiments on different settings: (1) combined scoring and ranking for unsupervised key-phrase extraction; (2) embedding representation; (3) clustering methods.

\subsection{Combined Scoring in Key-phrase Extraction (Q1)}

For key-phrase extraction, we compared combined scoring method with four base models. One is \textbf{TextRank}\footnote{Implemented on pke python library (\url{https://github.com/boudinfl/pke})} \cite{mihalcea-tarau-2004-textrank}, a graph-based keyword extraction module. The other two are SIFRank-ELMo, SIFRank-Bert and SIFRank-ConceptNet, where the difference lies in the underlying pre-trained word embedding representation. Our key-phrase extraction method is the extension of base models by combined scoring and ranking with two other scores. We optimized the scores' weights based on evaluation and set weights to $0.1$ for both domain relevance and phrase quality. Experimental results form table \ref{table:ensemble} show that combined scoring methods outperform their original base models in three settings (TextRank, SIFRank-ELMo and SIFRank-ConceptNet), where SIFRank-Bert only performs better than baseline in Top10 and Top15 key-phrases. Table \ref{table:scores} also shows positive effect when adding two scores to baselines. Meanwhile, the quality score shows larger impact than domain relevance. We think it is because domain relevance score is sensitive to the quality of domain glossaries. Also, good key-phrases in scientific literature usually contain similar structure, e.g., multi-word expression. It also indicates that filtering out 'poor' candidate phrases can largely contribute to better extracting performance. 

\begin{table}[htb]
\resizebox{0.5\textwidth}{!}{
\begin{tabular}{|c|c|c|c|c|}
\hline
\multicolumn{1}{|l|}{}              &                               & \multicolumn{3}{c|}{Top10} \\ \hline
\multirow{4}{*}{SIFRank-ELMo}       & \multicolumn{1}{l|}{Baseline} & 0.4327  & 0.3866  & 0.4083 \\ \cline{2-5} 
                                    & + Domain Relevance            & 0.4446  & 0.3972  & 0.4195 \\ \cline{2-5} 
                                    & + Phrase Quality              & 0.4809  & 0.4297  & 0.4539 \\ \cline{2-5} 
                                    & Combined Scoring              & 0.4834  & 0.4319  & 0.4562 \\ \hline
\multirow{4}{*}{SIFRank-Bert}       & \multicolumn{1}{l|}{Baseline} & 0.4418  & 0.3947  & 0.4169 \\ \cline{2-5} 
                                    & + Domain Relevance            & 0.4372  & 0.3906  & 0.4126 \\ \cline{2-5} 
                                    & + Phrase Quality              & 0.4425  & 0.3953  & 0.4176 \\ \cline{2-5} 
                                    & Combined Scoring              & 0.4442  & 0.3969  & 0.4192 \\ \hline
\multirow{4}{*}{SIFRank-ConceptNet} & \multicolumn{1}{l|}{Baseline} & 0.4257  & 0.3803  & 0.4017 \\ \cline{2-5} 
                                    & + Domain Relevance            & 0.4404  & 0.3934  & 0.4156 \\ \cline{2-5} 
                                    & + Phrase Quality              & 0.4823  & 0.4309  & 0.4552 \\ \cline{2-5} 
                                    & Combined Scoring              & 0.4774  & 0.4266  & 0.4103 \\ \hline
\end{tabular}}
\caption{Comparison of the impact of three different scores to key-phrase performance.}\label{table:scores}
\end{table}

\begin{figure}[htb]
    \centering
    \includegraphics[width=0.5\textwidth]{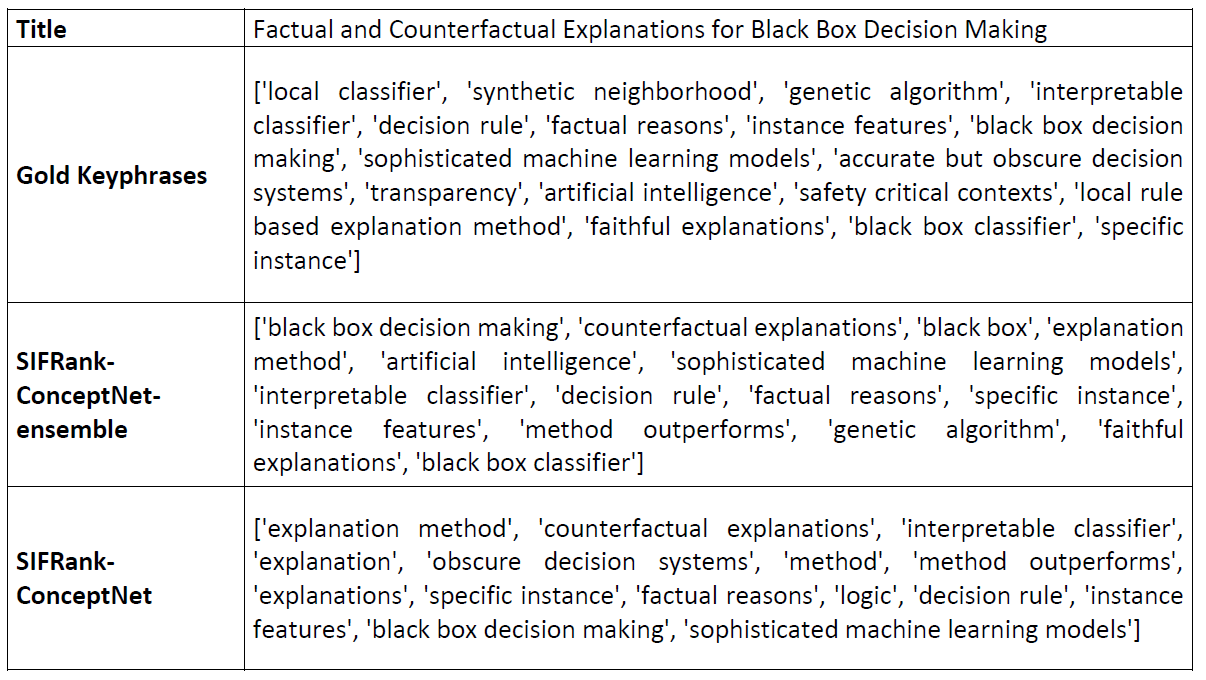}
    \caption{Example from top-15 extracted key-phrases.}
    \label{fig:kpexample}
\end{figure}

From the example of top-15 extracted key-phrases (in Fig. \ref{fig:kpexample}, adding domain relevance and phrase quality could reduce the rank of uni-grams (\textit{``method"}, \textit{``logic"}, \textit{``explanation"}) as well as terms with abstract meanings (\textit{``explanation method"}). However, it still has limitation on nested key-phrases with similar meanings (\textit{``black box decision making"} and \textit{``black box"}) and wrong candidates from selection (\textit{``method outperforms"}).

\subsection{Word Embedding (Q2)}

Pre-trained embeddings are utilized for sentence and phrase representation in our method. For ConceptNet embedding, each phrase is segmented by the longest matching terms in the embedding index and encoded by average embedding vectors. Since ELMo encodes phrases token by token, we take the mean vector of all tokens in the phrase. Comparing the three settings of pre-trained word embedding used in SIFRank model, both SIFRank-ELMo and SIFRank-ConceptNet based models present similar performance, while SIFRank-ELMo has slightly higher (Table \ref{table:ensemble}). 

\begin{table}[htb]
\centering
\begin{tabular}{|c|c|}
\hline
                            & Time \\ \hline
SIFRank-ELMo      & 1650.81                           \\ \hline
SIFRank-ConceptNet & 258.13                           \\ \hline
\end{tabular}
\caption{Execution time (in seconds) of two key-phrase extraction methods, including loading embedding.}\label{table:time}
\end{table}


However, contextualized models as ELMo and Bert require much more execution time than ConceptNet (Table \ref{table:time}). Here it is worth noting that ELMo and Bert generate embeddings from large natural language text corpus, while ConceptNet embeddings are generated from semantic network. However, our previous key-phrase extraction results do not show a large difference between ELMo based and ConceptNet based methods. Therefore, the NumberBatch embeddings based on ConceptNet are more efficient for short term extraction. 

\subsection{Clustering (Q3)}

In the clustering module, each key-phrase is treated as an independent ontological concept term. Term-level clustering group terms together based on cosine similarity of embedding from ConceptNet NumberBatch. Spherical $k$-means and hierarchical agglomerative clustering (HAC) are evaluated in our clustering module. HAC uses average linkage and cosine distance. For clustering experiments on the {\em XAI dataset}, terms are selected from the best model in key-phrase extraction experiment, with key-phrases post-processing and cleaning discussed above. 

\begin{table*}[htb]
\centering
\resizebox{\textwidth}{!}{
\begin{tabular}{|cccc|}
\hline
1                             & 2                                    & 3                                & 4                                          \\ \hline
visual\_analytics             & object\_detection                    & white\_box                       & fuzzy\_system                              \\ 
visual\_analytics\_workflow   & object\_detection\_system            & white\_box\_solution             & hierarchical\_fuzzy\_system                \\ 
visual\_analytics\_tool       & object\_detection\_framework         & white\_box\_method               & fuzzy\_system\_complexity                  \\ 
visual\_analytics\_framework  & interpretable\_object\_detection     & black\_box\_decision\_making     & evolutionary\_fuzzy\_system                \\
visual\_analytics\_researcher & robust\_object\_detection            & equivalent\_white\_box\_solution & neuro\_fuzzy\_system                       \\ 
visual\_analytics\_paradigm   & occlusion\_robust\_object\_detection & black\_box\_nature               & interpretable\_fuzzy\_system               \\ 
visual\_analytics\_solution   & semantic\_object\_part\_detector     & black\_box\_prediction           & fuzzy\_method \\ \hline
\end{tabular}}

\caption{Example of cluster-wise results on Spherical $k$-means clustering of XAI publications dataset.}\label{table:spkm_xai_example}
\end{table*}

\begin{figure}[htb]
    \centering
      \includegraphics[width=0.4\textwidth]{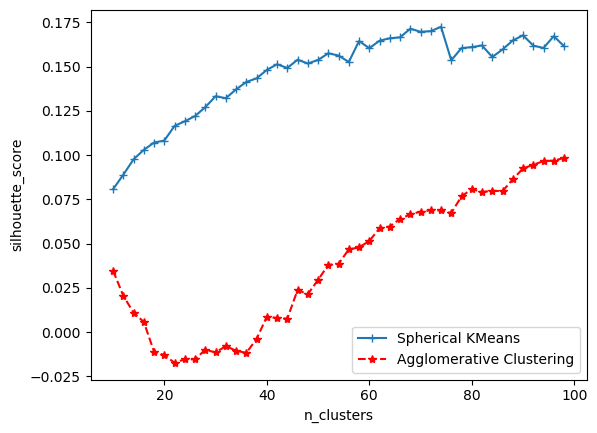}
      \caption{Results of silhouette coefficient score with \textit{n} clusters.}
      \label{fig:sil}
\end{figure}

Silhouette score in Figure \ref{fig:sil} shows that the curve of agglomerative clustering does not reach a peak within range of 100 clusters, while Spherical $k$-means reach its highest score at 89 clusters. Also, spherical $k$-means gets better cluster quality than hierarchical agglomerative clustering, which is also proved by results in Table \ref{table:clus_xai}.

\begin{table}[htb]
\centering
\begin{tabular}{|c|c|}
\hline
              & silhouette score  \\ \hline
spherical $k$-means & 0.1615    \\ \hline
  HAC       & 0.0690     \\ \hline
\end{tabular}
\caption{Clustering analysis on XAI publications dataset. Number of clusters is set to 74.}\label{table:clus_xai}
\end{table}

\begin{figure}[htb]
    \centering
    \includegraphics[width=0.45\textwidth]{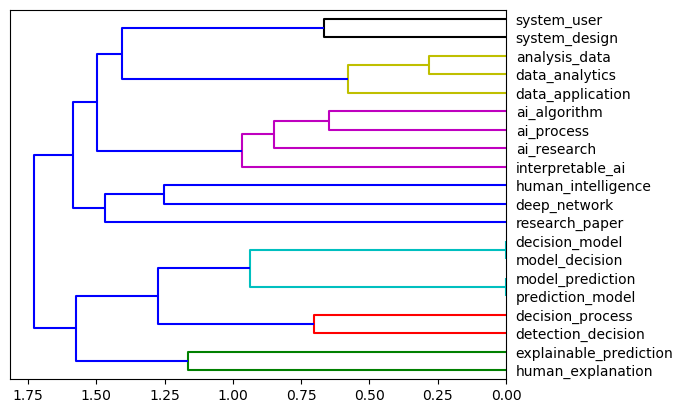}

    \caption{Example of Hierarchical Agglomerative Clustering Dendrogram.}
    \label{fig:hac_example}
\end{figure}

Theoretically, silhouette score ranges from -1 to 1, where 1 indicates better separation among clusters and 0 means overlapping between clusters. Even though both clustering algorithms do not reach highly significant silhouette score, analysis of clusters output proves semantic coherence of terms within clusters (Fig. \ref{fig:hac_example} and Table \ref{table:spkm_xai_example}), which can be identified as semantic categories of these key-phrases. Table \ref{table:spkm_xai_example} selects four example clusters, where terms in table are ranked by its distance to its cluster center. Clusters in the Table shows categories of \textit{``visual analytic"} (cluster 1), \textit{``object detection"} (cluster 2), \textit{``white box"} (cluster 3) and \textit{``fuzzy system"} (cluster 4). By manually analyzing created clusters, some observation can be made that:

 \begin{itemize}
    \item Terms within one cluster show high similarity in sub-words, while sometimes the same sub-words indicate semantic relatedness.
    \item Central meaning ``word" represents the topic or category found in the cluster, which further determines whether it can be used as a part of classification scheme.
 \end{itemize}
 
Possible reasons for such results could be due to: (1) we may encounter a limitation in regard to the embedding representation of terms. Fine-tuning ConceptNet embeddings would require a network of domain-specific ontologies; thus, it is not applicable in our research. Pre-trained embedding may have limited discriminative power in a specific domain; (2) a limitation is considered concerning the clustering algorithms. Generic clustering algorithms assume data points can be separated. Internal evaluations also measure the separation of clusters. We notice that clusters overlap in the embedding space; thus, clustering may not be able to separate clusters.

\section{Conclusion}

This paper proposes a joint framework of unsupervised key-phrase extraction and semantic term clustering to automate systematic mapping studies. Experiments are conducted using publications from the domain of \textit{Explanable Artificial Intelligence (XAI)}". In detail, we examined the ensemble ranking scores, ConceptNet word embedding, and clustering performance. 

Results in key-phrase extraction demonstrate the effectiveness of ensemble ranking scores from different perspectives, where domain knowledge (in terms of glossaries and domain corpus) finds highly relevant terms which can be further considered as constraints and external resources for weak supervision. ConceptNet based word embedding performs as well as contextualized word embeddings, with much less execution time. Findings are further useful to guide the choice of a suitable word embedding method in terms of tasks and use cases. Semantic term clustering can group semantically similar terms within clusters, still we suggest some minimal human involvement may help refine and select high-quality keywords clusters based on use cases, with the bulk of the work been primarily performed by the algorithm. 

Above all, we hope our research can give a new perspective of automating keywording and classification scheme steps in systematic mapping studies towards faster and more convenient solutions in an open research knowledge era. For future work, the role of human involvement can be further evaluated with having specific use cases in mind. Finally, ontology-related techniques could be explored as means to refining keywords.

\bibliography{main.bib}
\bibliographystyle{aaai}

\end{document}